\pdfoutput=1

\documentclass[11pt]{article}

\usepackage[final]{acl}
\usepackage{times}
\usepackage{latexsym}

\usepackage[T1]{fontenc}

\usepackage[utf8]{inputenc}
\usepackage{booktabs,multirow,rotating,bigstrut}
\usepackage{siunitx}
\usepackage{colortbl}
\usepackage{tabularx}
\usepackage{amssymb}
\usepackage{pifont}

\usepackage{microtype}
\usepackage{hyperref}
\makeatletter
\def\UrlAlphabet{%
      \do\a\do\b\do\c\do\d\do\e\do\f\do\g\do\h\do\i\do\j%
      \do\k\do\l\do\m\do\n\do\o\do\p\do\q\do\r\do\s\do\t%
      \do\u\do\v\do\w\do\x\do\y\do\z\do\A\do\B\do\C\do\D%
      \do\E\do\F\do\G\do\H\do\I\do\J\do\K\do\L\do\M\do\N%
      \do\O\do\P\do\Q\do\R\do\S\do\T\do\U\do\V\do\W\do\X%
      \do\Y\do\Z}
\def\UrlDigits{\do\1\do\2\do\3\do\4\do\5\do\6\do\7\do\8\do\9\do\0}
\g@addto@macro{\UrlBreaks}{\UrlOrds}
\g@addto@macro{\UrlBreaks}{\UrlAlphabet}
\g@addto@macro{\UrlBreaks}{\UrlDigits}

\usepackage{inconsolata}
\usepackage{amsmath}

\title{BIDER: Bridging Knowledge Inconsistency for Efficient Retrieval-Augmented LLMs via Key Supporting Evidence}
\author{Jiajie Jin \quad Yutao Zhu \quad Yujia Zhou \quad Zhicheng Dou\thanks{*Corresponding author}\ \\
Gaoling School of Artificial Intelligence, Renmin University of China \\
\texttt{\{jinjiajie,zhouyujia,dou\}@ruc.edu.cn, yutaozhu94@gmail.com}
} 

\begin{document}
\maketitle

\begin{abstract}

Retrieval-augmented large language models (LLMs) have demonstrated efficacy in knowledge-intensive tasks such as open-domain QA, addressing inherent challenges in knowledge update and factual inadequacy.
However, inconsistencies between retrieval knowledge and the necessary knowledge for LLMs, leading to a decline in LLM's answer quality.
This paper introduces BIDER, an approach that refines retrieval documents into Key Supporting Evidence (KSE) through knowledge synthesis, supervised fine-tuning (SFT), and preference alignment. 
We train BIDER by learning from crafting KSE, while maximizing its output to align with LLM's information acquisition preferences through reinforcement learning.  
Evaluations across five datasets show BIDER boosts LLMs' answer quality by 7\% while reducing input content length in retrieval documents by 80\%, outperforming existing methods. The proposed KSE simulation effectively equips LLMs with essential information for accurate question answering.

\end{abstract}
\section{Introduction}

Large language models (LLMs) are currently developing rapidly and showing tremendous capabilities~\cite{achiam2023gpt,touvron2023llama}. 
Nevertheless, they face challenges in knowledge updates and furnishing factual responses
~\cite{bang2023hallucination},
especially in knowledge-intensive tasks like open-domain QA~\cite{jiang-etal-2023-active}.
To address these issues, retrieval-augmented generation (RAG) has emerged as a promising approach~\cite{lewis2020retrieval,guu2020realm}.
Retrieval-augmented methodologies serve to mitigate the drawbacks of LLMs by incorporating external knowledge, thereby enhancing the quality and reliability of generated answers~\cite{izacard2022atlas,shi2023replug,press-etal-2023-measuring}. 

The standard RAG procedure involves retrieving pertinent documents related to a given question and subsequently inputting these documents as auxiliary information directly into the prompt. This strategic utilization enables the model to capitalize on its advanced text comprehension skills, facilitating the generation of precise and contextually appropriate answers.
\begin{figure}[!t]
    \centering
    \setlength{\abovecaptionskip}{0.1cm}
    \setlength{\belowcaptionskip}{-0.1cm}
    \includegraphics[width=0.8\linewidth]{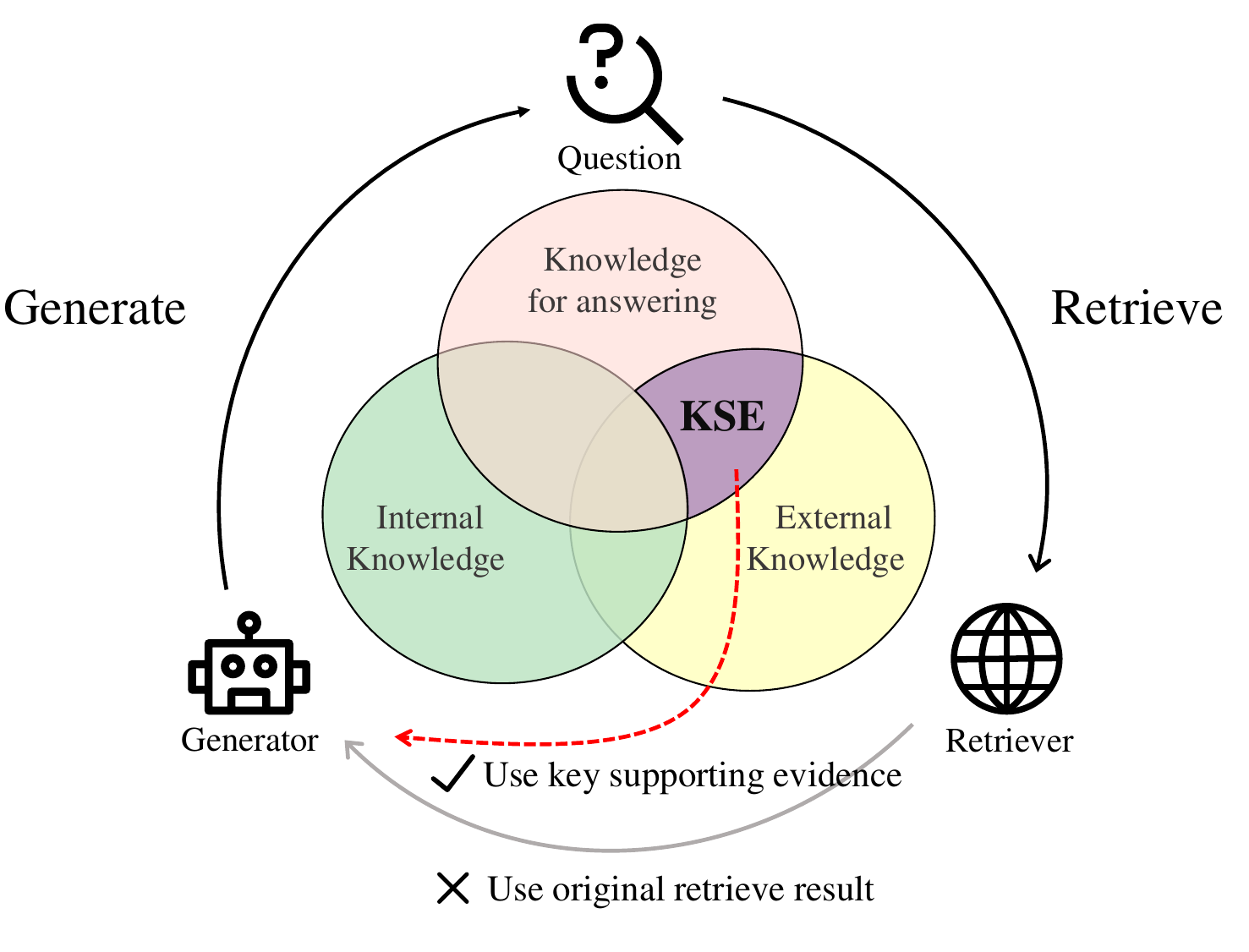}
    \caption{Key supporting evidence in RAG framework.}
    \vspace{-3mm}
    \label{fig:intro}
\end{figure}

However, retrieval-augmented LLMs are not always beneficial. Due to imperfections in the retrieval system and the inaccessibility of LLM's self-knowledge~\cite{wang-etal-2023-self-knowledge}, the retrieved documents provided to LLM are frequently lengthy and noisy, which can detrimentally affect generation quality~\cite{howaffects,shi2023irrelevant}.

Recognizing this decline, recent researches have made strides in optimizing retrieved documents. These efforts aim to mitigate noise in retrieved documents by employing sorting mechanisms to retain the most pertinent sentences~\cite{xu2023recomp,arefeen2023leancontext}, summarizing the retrieved text~\cite{xu2023recomp}, and eliminating content that contributes minimally to the model's understanding~\cite{li2023selectivecontext,jiang-etal-2023-longllmlingua} or hinders effective generation~\cite{yang-etal-2023-prca}. 

While prior methods have shown progress in enhancing the quality of retrieved documents, they often rely excessively on feedback from the generator, overlooking the essential knowledge required for addressing the questions themselves. 
This over-reliance on LLM's feedback is not only insufficient but also susceptible to the instability of LLM feedback, potentially resulting in the loss of crucial information and the retention of noisy elements.
We argue that this limitation might stem from the neglect of knowledge inconsistency between the retrieved results and the knowledge truly required by the model for answering the question. We term this essential knowledge as Key Supporting Evidence (KSE).
As shown in Figure~\ref{fig:intro}, due to the imperfections of the retrieval system and the inaccessibility of LLM's self-knowledge~\cite{wang-etal-2023-self-knowledge}, retrieved results often contain numerous noise elements beyond key supporting evidence.

\begin{figure*}[!t]
    \centering
    \setlength{\abovecaptionskip}{0.1cm}
    \setlength{\belowcaptionskip}{-0.1cm}
    \includegraphics[width=1.0\linewidth]{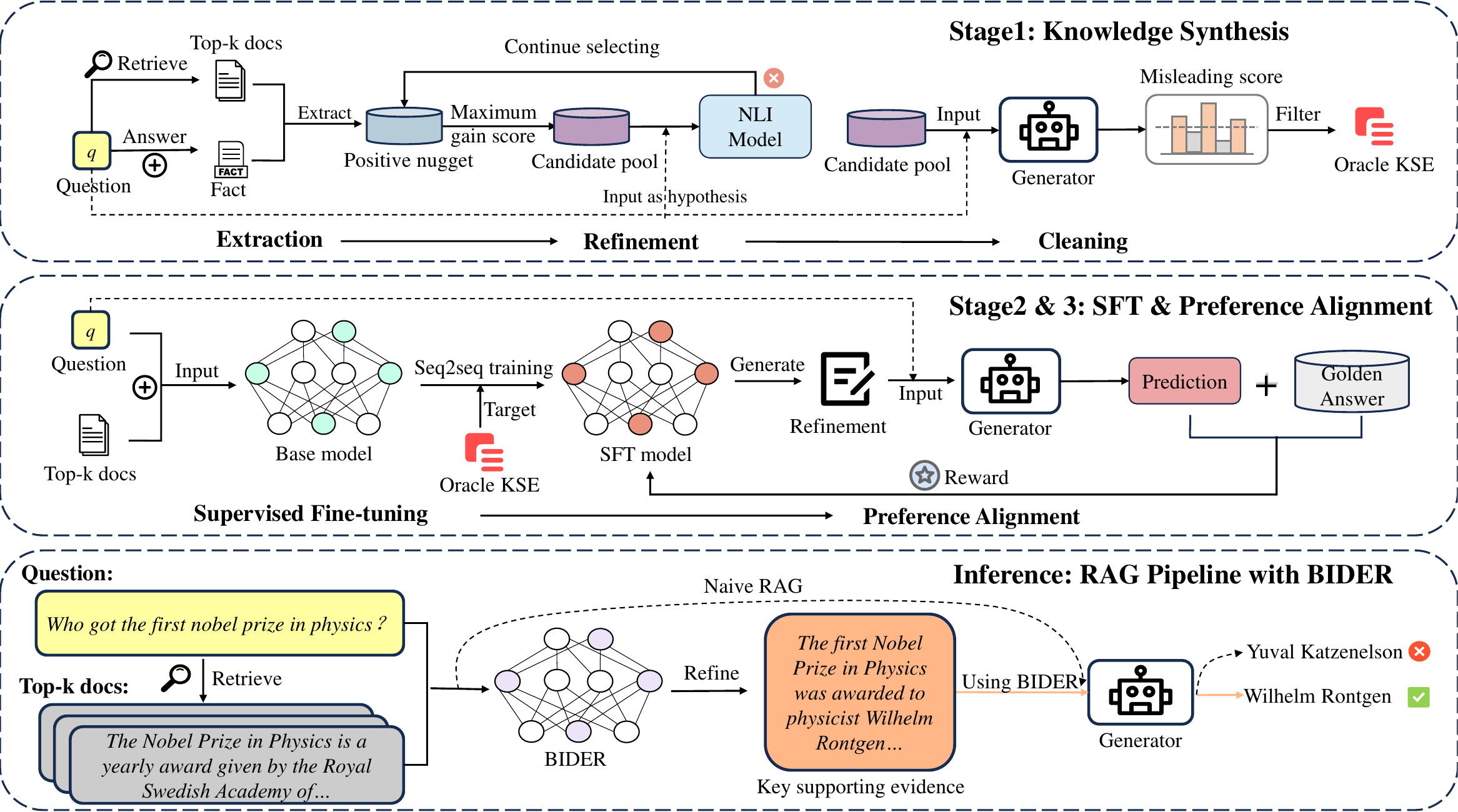}
    \caption{The overall architecture of BIDER. The first two lines represent the training process, which consists of three stages, and the last line represents the inference process of RAG with BIDER.}
    \label{fig:main_method}
\end{figure*}

To address the aforementioned knowledge inconsistency issue, we propose BIDER(BrIDging knowledge inconsistency for efficient Retrieval-augmented LLMs), a method designed to refine retrieval documents into KSE. The overall training process of BIDER consists of three stages, integrating the strengths of both supervised and reinforcement learning, as shown in Figure~\ref{fig:main_method}.
In the knowledge synthesizing stage, we employ a meticulous three-step process to synthesize authentic KSE. In the supervised fine-tuning stage, we construct a seq2seq model to learn the mapping from retrieved documents to KSE. Finally, in the preference alignment stage, we leverage reinforcement learning techniques to align the developed model with the preferences of the downstream LLM. This alignment ensures that the refined retrieval documents contain coherent and easily digestible key information, which is crucial for the LLM to generate accurate and informative responses.

We evaluate the effectiveness of our method on five datasets from three types of knowledge-intensive tasks, i.e., NQ, TQA, and HotpotQA for open-domain QA, WoW for dialogue generation, and FEVER for fact verification. Results show that our method achieves better generation performance while reducing the input information length by 80\%, effectively condensing retrieved documents, and outperforming existing methods. 
We also validate the advantages of our proposed KSE data construction process and investigate the impact of the preference alignment stage on the final results.
Furthermore, we validate the robustness of our approach under various text retrieval quality conditions.

The main contributions of this work are:
(1) We propose a three-step knowledge synthesis method to generate oracle KSE.
(2) We introduce a method to refine retrieval documents into KSE, thereby bridging knowledge inconsistencies between retrieval documents and the knowledge required by LLMs for answering. 
(3) We train the refiner model using supervised distillation and preference alignment techniques, efficiently enhancing RAG performance during inference by reducing input length and improving answer quality.

\section{Related Work}
\subsection{RAG for LLMs}

In knowledge-intensive tasks~\cite{petroni-etal-2021-kilt}, RAG ~\cite{lewis2020retrieval} has been introduced to enhance generative outcomes by incorporating external knowledge sources. 
In previous work, the retriever and generator are usually jointly trained end-to-end~\cite{guu2020realm,lewis2020retrieval,borgeaud2022retro}. With the advent of LLMs~\cite{achiam2023gpt,touvron2023llama}, most works now directly use them as generators due to their strong text comprehension ability, without the need for additional training~\cite{jiang-etal-2023-active,yao2023react,shinn2023reflexion}. 
While this approach demonstrates efficiency, it introduces new challenges, including susceptibility to interference from irrelevant content~\cite{shi2023irrelevant,bai2023longbench,mallen2023llm_memorization}, insufficient attention to middle positions~\cite{liu2023lost}, and increased inference costs~\cite{NEURIPS2022_gpt3int}. 
Our method refines retrieved documents to eliminate noise, significantly reducing the input required for inference. By learning information retrieval preferences from LLM feedback, it provides the LLM with text that is more informative and easily captures relevant information, offering a substantial solution to the aforementioned issues.

\subsection{Knowledge Refinement for RAG}

Recent works leverage the capabilities of LLMs to identify pertinent information from various perspectives.
Some approaches directly task the LLM with summarizing retrieval documents to identify pertinent information~\cite{laskar2023cqsumdp,chen2023walking, gilbert2023semantic,xu2023recomp}.
Moreover, certain methods employ smaller models to calculate perplexity as an importance indicator for filtering low-information text~\cite{li2023selectivecontext,jiang-etal-2023-longllmlingua}. \citet{xu2023recomp} employ the LLM to assess the utility of each sentence in retrieval documents, using this information as labels to train a ranking model.
Other works leverage LLM feedback for training; for instance, \citet{arefeen2023leancontext} train a ranking model using reinforcement learning to retain top-ranked sentences, and \citet{yang-etal-2023-prca} design a reward mechanism to train a refinement model for retrieval documents. While these methods are effective, they are constrained by the instability of LLM feedback, providing limited guidance on specific information deemed valuable, which results in inefficient and suboptimal training outcomes. In contrast, our method employs well-designed key supporting evidence as a training objective, allowing the refiner to learn knowledge comprehensively before reinforcement learning, ensuring the provision of knowledge that better aligns with the LLM's needs.

\section{BIDER: a Knowledge Refiner for RAG}
Our objective is to furnish the necessary knowledge for a generator, specifically the KSE as defined earlier, to answer a question. Since authentic KSE is unattainable, we employ a synthesize-and-learn paradigm. 
We design a method for synthesizing authentic KSE, training the refiner to learn the map from retrieval documents to constructed KSE, and adapting the model's information acquisition preferences based on the generator's feedback.

The overall framework of BIDER is illustrated in Figure~\ref{fig:main_method}.
In this section, we first formulate the research problem (\S\ref{sec:s1}). Then, we introduce the details of three training stages of BIDER, including Knowledge Synthesis (\S\ref{sec:s2}),  Supervised Distillation (\S\ref{sec:s3}), and Preference Alignment (\S\ref{sec:s4}).

\subsection{Problem Formulation}
\label{sec:s1}

In this problem, we assume that a document collection $\mathcal{C}$, a fixed retriever $\mathcal{R}$, and a fixed generator $\mathcal{G}$ are provided. For a given question $q$ and its corresponding golden answer $o$, we assume that $K$ documents are retrieved by retriever $\mathcal{R}$, denoted as $\mathcal{D}_q=\{{d_i}\}_{i=1}^K$. In the naive RAG framework, $\mathcal{D}_q$ is directly incorporated into the generator's input to obtain the output answer. We aim to find the optimal mapping function $\mathcal{F}^*$ for the retrieved documents $\mathcal{D}_q$, in order for the generator $\mathcal{G}$ to use $\mathcal{F}^*(\mathcal{D}_q)$ and achieve the best output results. We design BIDER to act as the mapping function, refining retrieval documents to make them more suitable for the input preferences of the generator.

\subsection{Knowledge Synthesis Stage}
\label{sec:s2}

We design a three-step method to gradually synthesize authentic KSE.
 
\textbf{(1) Nugget Extraction.} We initially narrow down the scope of knowledge helpful for answering by extracting nuggets from the retrieval documents. Here a nugget can be a sentence, a passage, or even a key phrase. In this paper, we use sentences as nuggets because using sentences already yields robust and consistent results. We will explore approaches with different nugget granularities in our future work. 
For each input question $q$ and its corresponding golden answer $o$, we first formulate them into a fact $f = \text{concat}(q,o)$ to ensure comprehensive semantic representation.\footnote{For FEVER and HotpotQA where the answers have no actual semantic meaning, we use annotated evidence as a golden answer to ensure more accurate semantic information.}
Then, we use $f$ as the query to perform sentence-level nugget retrieval in the retrieved documents $\mathcal{D}_q$ to remove noise and retain helpful sentences.
In nugget retrieval, $\mathcal{D}_q$ is split into nuggets and transformed into vectors, while the query is vectorized. 
Based on the similarity between the query vector and nugget vectors, we obtain a positive nugget set $\mathcal{S}$ including retrieved top $K$ nuggets:
\begin{equation}
\mathcal{S} = \mathop{\text{TopK}}\limits_{s\in \mathcal{D}_q}(\text{sim}(s, f)).
\end{equation}
Here, $\text{sim}(\cdot,\cdot)$ represents the function for calculating semantic similarity by the E5 model~\cite{wang2022e5}, and $K$ is a hyperparameter. A larger $K$ can improve the recall of relevant information in $\mathcal{D}_q$, but it also raises the risk of including more irrelevant information. 

\textbf{(2) Nugget Refinement.} While the extraction step effectively reduces noise in retrieved documents, there may be redundancy in $\mathcal{S}$.
Therefore, we further design an iterative selection method to retain the minimal nugget subset necessary for answering the question.

Initially, we set up a candidate pool $\mathcal{P}$. 
In each round, we calculate a gain score for each nugget in $\mathcal{S}$, which represents the degree of assistance in answering the question. The gain score is defined as follows:
\begin{equation}
\kappa_i = \text{sim}(s_i, f) - \frac{1}{|\mathcal{P}|}\sum_{s_j\in \mathcal{P}}\text{sim}(s_i,s_j).
\end{equation}
This takes into account the importance of the nugget itself as well as its duplication with the already-selected nuggets.
Then, we select the sentence with the highest $\kappa_i$ from $\mathcal{S}$ and move it to the candidate pool $\mathcal{P}$. After moving, we use an NLI model to measure to what extent the candidate pool $\mathcal{P}$ can support answering $q$, i.e., yielding a support degree $\eta_k$ in $k$-th nugget selection. 

The iterative selecting process will terminate in two cases: (1) when the support degree $\eta_k$ exceeds $\lambda_{\text{max}}$; (2)when the difference in support degree between two rounds, $\eta_k - \eta_{k-1}$, is less than $\lambda_{\text{min}}$, where $\lambda_{\text{max}}$ and $\lambda_{\text{min}}$ are predefined thresholds.
This aims to avoid introducing redundant information, especially in scenarios where retrieval documents fail to furnish adequate information, such as instances where the retriever's quality is subpar or when the posed question is challenging.

\textbf{(3) Nugget Cleaning.} The candidate pool $\mathcal{P}$ from the previous stage serves as the minimal subset of information necessary for answering. However, we have yet to consider the knowledge intrinsic to the generator itself, which encompasses information either known by LLM or detrimental to its generation.
To mitigate conflicts arising from the disparity between external and internal knowledge, we conduct nugget cleaning in the candidate pool.
In our experiments, we observe that the first nugget within the candidate pool is usually important. 
To avoid unintentionally removing vital information, we retain the first nugget directly and perform nugget cleaning for the left set. 

For each nugget $s_i (i\geq 2)$ in $\mathcal{P}$, we assess its influence on the generator by determining whether it contributes to the model's output improvement when utilized as input. Specifically, we calculate the change in the log probability of generating the correct answer $o$ between the model's output before and after the inclusion of the nugget. This score for each nugget $s_i$ is denoted as 
\begin{equation}
\tau_i^{\prime} =  \log \frac{\mathcal{G}(o|q\oplus s_i)}{\mathcal{G}(o|q)}, i\geq 2. 
\end{equation}
where $\mathcal{G}$ represents the generator.

Subsequently, we normalize all scores within the candidate pool to derive the final score $\tau_i$:
\begin{equation}
\tau_i = \frac{\tau_i^{\prime}}{\sum_{j=2}^{|\mathcal{P}|} \tau_j^{\prime}}, i\geq 2.
\end{equation}
Nuggets with scores below $\lambda_{\text{lm}}$ are deemed unhelpful or potentially detrimental to the generator's response and are consequently excluded from the candidate pool. \textbf{The surviving nuggets in the filtered candidate pool represent the ultimate oracle KSE}, which correspond to the distillation results for each sample triplet $(q,o,\mathcal{D}_q)$.

\subsection{Supervised Distillation Stage}
\label{sec:s3}

In this stage, we aim to develop BIDER to acquire the ability to comprehend the relationship between retrieval documents and oracle KSE. This enhancement will enable BIDER to effectively refine its output during inference, particularly when provided only with the question.

A common approach is to consider this as a ranking task ~\citep{xu2023recomp,liu2019bertsum}, using the nuggets extracted in the previous section as positive examples and other nuggets as negative examples for training the ranker. Although this method can relatively stably filter information, it is not able to effectively generate content that can adapt to the input of the generation model, as the refined content can only come from the original text. 

We model the task as a seq2seq task, which is similar to the idea of pointer network ~\citep{see-etal-2017-get, gu2016copynet}. This method ensures the flexibility of refinement while enhancing the potential of the generation model in expression. Meanwhile, this serialization modeling approach makes it easier for the model to capture the generated sentences during generation. In Section~\ref{sec:exp_results}, we will compare the two methods and demonstrate the effectiveness of our approach.

We use a pre-trained seq2seq model as the backbone model. For each sample triplet $(q, o, \mathcal{D}_q)$, the refiner model's input is the concatenation of the question and the original retrieval document: $q \oplus \mathcal{D}_q$. For ease of processing, we add separators between each document in $\mathcal{D}_q$ and merge them into one string. The target output of the model is the $\mathcal{P}$ extracted in the Knowledge Synthesis Stage, where each nugget is merged into one string in order. The training loss function of the model is the cross-entropy loss between the model output and the target output.

\subsection{Preference Alignment Stage}
\label{sec:s4}

Inspired by the RLHF technology~\citep{Ziegler2019FineTuningLM,NEURIPS2020_summarizehf,NEURIPS2022_instruct}, we further enhance the adaptability of BIDER by incorporating feedback from a downstream LLM.

Specifically, we model the optimization problem of the model as a RL problem, with the objective to generate content that conforms to the LLM's information acquisition preferences without losing its original information capturing ability. The refiner model 
$\mathcal{M}$ to be optimized acts as a policy, where its action space encompasses all tokens in the vocabulary. We use the CLIP version of the PPO algorithm~\citep{Schulman2017PPO} for optimization, which uses CLIP to control the magnitude of model updates.
The loss function consists of three parts:
\begin{equation}
L_t^{\text{ALL}} = \mathrm{E}_t[L_t^{\text{CLIP}} - L_t^{\text{VF}}+ L_t^{\text{BONUS}}].    
\end{equation}
$L_t^{\text{CLIP}}$ is the primary objective function for optimizing the policy at step $t$, expressed as:
\begin{gather}
    L_t^{\text{CLIP}} = \text{min}(r_t\hat{A_t}, \text{clip}(r_t, 1-\epsilon,1+\epsilon)\hat{A_t} \\
    r_t=\frac{\pi_\theta(y|x)}{\pi_\text{old}(y|x)},
\end{gather}
where $\epsilon$ is a hyperparameter to control the policy update magnitude, $r_t$ represents the conditional generation probability ratio between the new policy and the old policy, and $A_t$ denotes the estimated value of the advantage function at step $t$, calculated from Generalized Advantage Estimation (GAE)~\citep{GAE}:
$$\hat{A_t} = \sum_{l=0}^{T-t+1}(\gamma \lambda)^l \left(R_t + \gamma V(s_{t+1}) - V\left(s_t\right)\right),$$
where $\gamma$ and $\lambda$ are hyperparameters. $V$ represents the critic network used to estimate expected rewards, and $R_t$ indicates the reward at step $t$.
$L_t^{VF}(\theta)$ is the squared error between the predicted reward and the actual reward output by the critic network, used to fit the critic network:
\begin{equation}
   L_T^{\text{VF}}(\theta) = (V_\theta(s_t)-R_t)^2. 
\end{equation}
$L_t^{\text{VF}}(\theta)$ is an entropy bonus designed to ensure the model can explore sufficiently.

To calculate the above loss, we need a well-defined reward function $R_t$. Considering that the downstream LLM is highly sensitive to the overall information density and the position of key information, providing rewards before all sentences are generated could lead to inaccurate guidance. Thus, we design a segmented reward function: 
\begin{equation*} 
R_t = 
    \begin{cases} 
    0, & s_t \neq \langle \text{EOF} \rangle, \\ 
    \text{F}_1(a_{\text{pred}},o) - \text{F}_1(a_{\text{ori}},o), & s_t = \langle \text{EOF} \rangle. 
    \end{cases}
\end{equation*}
where $a_{\text{pred}}$ and $a_{\text{ori}}$ represent the answers generated by LLM based on the refiner result and original retrieval result respectively
, $\langle \text{EOF} \rangle$ represents the end-of-sentence symbol. We generate answers from the LLM using the original document and refined results separately as references, and evaluate the quality of the refiner's distillation of the retrieved document by comparing the token-level $\text{F}_1$ scores of these two types of answers with the golden answer.

\begin{table*}[!t]
  \centering
   \resizebox{\linewidth}{!}
   {
    \begin{tabular}{lcccccccccccc}
    \toprule
    \multirow{2}[3]{*}{\textbf{Methods}} & \multicolumn{2}{c}{\textbf{NQ}} & \multicolumn{2}{c}{\textbf{TQA}} & \multicolumn{2}{c}{\textbf{Fever}} & \multicolumn{2}{c}{\textbf{HotPotQA}} & \multicolumn{2}{c}{\textbf{Wow}} & \multirow{2}[3]{*}{\textbf{Avg}} & \multirow{2}[3]{*}{\textbf{Avg tok}} \\
\cmidrule(lr){2-3}\cmidrule(lr){4-5}\cmidrule(lr){6-7}\cmidrule(lr){8-9}\cmidrule(lr){10-11}         & \textbf{EM} & \textbf{\# tok} & \textbf{EM} & \textbf{\# tok} & \textbf{Acc} & \textbf{\# tok} & \textbf{F1} & \textbf{\# tok} & \textbf{F1} & \textbf{\# tok} &       &  \\
    \midrule
    \multicolumn{13}{l}{\textit{\textbf{Without refinement}}} \\
    \midrule
    Original Prompt & 0.356 & 725   & 0.480 & 787   & 0.517 & 805   & 0.376 & 770   & 0.086 & 710   & 0.363 & 759 \\
    Zero-shot & 0.189 & 0     & 0.456 & 0     & 0.517 & 0     & 0.268 & 0     & 0.085 & 0     & 0.303 & 0 \\
    \midrule
    \multicolumn{13}{l}{\textit{\textbf{Extractive refinement}}} \\
    \midrule
    BM25  & 0.295 & 163   & 0.479 & 181   & 0.520 & 193   & 0.356 & 186   & 0.085 & 177   & 0.347 & 180 \\
    SBERT & 0.339 & 162   & \underline{0.512} & 183   & 0.521 & 192   & 0.36  & 187   & 0.086 & 178   & 0.364 & 180 \\
    LLM-Embedder & 0.357 & 161   & 0.503 & 179   & 0.522 & 192   & 0.352 & 186   & \underline{0.118} & 179   & 0.370 & 179 \\
    $\text{Bge-Reranker}^\ast$ & \underline{0.380} & 164   & 0.504 & 181   & 0.522 & 194   & \underline{0.384} & 186   & 0.117 & 180   & \underline{0.381} & 181 \\
    \midrule
    \multicolumn{13}{l}{\textit{\textbf{Abstractive refinement}}} \\
    \midrule
    BART-Summarizer & 0.326 & 185   & 0.507 & 204   & 0.518 & 215   & 0.369 & 254   & 0.085 & 194   & 0.361 & 210 \\
    Selective-Context & 0.263 & 203   & 0.439 & 225   & 0.522 & 236   & 0.332 & 234   & 0.081 & 220   & 0.327 & 224 \\
    LongLLMLingua & 0.221 & 251   & 0.433 & 175   & \textbf{0.551} & 111   & 0.302 & 222   & 0.077 & 124   & 0.317 & 177 \\
    BIDER(ours)  & \textbf{0.403} & 77   & \textbf{0.523} & 98   & \underline{0.524} & 93   & \textbf{0.386} & 113   & \textbf{0.122} & 69    & \textbf{0.390} & 90 \\
    \bottomrule
    \end{tabular}%
    }
    \caption{Evaluation results on five knowledge-intensive datasets.  The best results are in \textbf{bold} and second best results are \underline{underlined}. The method marked with $^\ast$ have undergone additional training.}
    \vspace{-1mm}
  \label{tab:main}
\end{table*}

\section{Experimental Setup}
\subsection{Datasets and Metrics}
\label{sec:data}

We experiment on five datasets of three knowledge-intensive tasks in the KILT benchmark \citep{petroni-etal-2021-kilt}: (1) \textbf{Open-domain QA}, including NaturalQuestions (NQ) \citep{kwiatkowski-etal-2019-natural}, TriviaQA (TQA) \cite{joshi-etal-2017-triviaqa},and HotpotQA \citep{yang-etal-2018-hotpotqa};
(2) \textbf{Dialog Generation}, including the Wizard of Wikipedia (WoW)\citep{dinan2018wizard}, where the generator is tasked with continuing the dialogue based on the preceding conversation history; 
(3) \textbf{Fact-checking}, including FEVER\citep{thorne-etal-2018-fever} that classifies a given claim as "SUPPORTS" or "REFUTES". 

\begin{table}[!t]
    \centering
    \small    
    \begin{tabular}{lccc}
        \toprule
        \textbf{Dataset} & \textbf{Task} & \textbf{Metric} & \textbf{\# Train / \#Test} \\
        \midrule
        NQ & Open-domain QA & EM & 79.1k / 2.6k \\
        TQA & Open-domain QA & EM & 78.7k / 11.3k \\
        HoPo & Open-domain QA & F1 & 88.8k / 5.6k \\
        WoW & Dialogue & F1 & 63.7k / 3.0k \\
        FEVER & Fact checking & Acc & 104.9k / 10.4k \\
        \bottomrule
    \end{tabular}    
    \caption{Statistics and task metrics for five datasets.}
    \label{tab:dataset}
\end{table}

We use Exact Match (EM) as the evaluation metric for NQ and TQA, use accuracy for FEVER, and use token-level $\mathrm{F}_1$~\citep{jiang-etal-2023-active} for HotpotQA and Wow.
Evaluation is conducted on top 1000 samples in the test set of NQ, TQA, and HotpotQA, while on the development set of FEVER and WoW.  Table~\ref{tab:dataset} provides detailed sample sizes and the evaluation metrics used for each dataset. 

\subsection{Baselines}
\label{sec:baseline}
We compare with two types of baselines.

\paragraph{Extractive Methods:} We employ three retrieval methods to extract sentences from retrieval documents, including BM25~\citep{xu2023recomp}, Sentence-BERT~\citep{reimers-2019-sentence-bert}, and LLM-Embedder~\citep{llm-embedder} which is trained with contrastive learning and feedback from LLM.
To demonstrate the superiority of modeling the task as a seq2seq problem in the supervised distillation stage, we fine-tuned the bge-reranker-large~\citep{bge_embedding} for comparison.

\paragraph{Abstractive Methods:} BART-Large is utilized for summarization~\citep{bart-cnn}, along with two state-of-the-art perplexity-based prompt refinement models: Selective Context~\citep{li2023selectivecontext} and LongLLMLingua~\citep{jiang-etal-2023-longllmlingua}.


\subsection{Implementation Details}
\label{sec:imp}

The size of the positive nugget set $K$ is set to 7. We utilize a T5-XXL model  as the NLI model with the threshold $\lambda_{\text{max}}$ set to 0.5, $\lambda_{\text{min}}$ set to 0.01,$\lambda_{\text{lm}}$ set to 0.05.\footnote{\url{https://huggingface.co/google/t5_xxl_true_nli_mixture}} We utilized BART-Large~\citep{bart-large} as the base model for BIDER. During training, we utilize AdamW~\citep{loshchilov2018adamw} as the optimizer with a learning rate of 5e-5, and a batch size of 32. In the preference alignment stage, $top_k$ is set to 10, and $top_p$ is set to 0.95. The training is implemented with HuggingFace Transformers~\citep{wolf-etal-2020-transformers} and PFRL~\citep{PFRL}. 
We use the December 2018 Wikipedia dump~\citep{karpukhin-etal-2020-dense,Lin_etal_SIGIR2021_Pyserini} as retrieval corpus, BM25 as retriever, and SimLM as reranker~\citep{wang-etal-2023-simlm} for the top 100 documents returned by the retriever.
LLAMA2-7B~\citep{touvron2023llama} is utilized as the generator to provide answers.


\section{Experimental Results}
\subsection{Main Results}\label{sec:exp_results}

Table~\ref{tab:main} reports the results of our approach alongside baseline methods on five knowledge-intensive datasets. It can be observed that 
our method outperforms the baseline on all datasets except FEVER, showcasing a notable performance advantage over other existing approaches, demonstrating around a 10\% performance increase on all datasets. We also observe a relatively small performance gap on the FEVER dataset, indicating a potential weakness in the model's ability to leverage retrieval documents for text-based verification tasks.
Compared to the original prompt, our method refines the retrieval documents to 20\% of their original length, achieving an average improvement of approximately 8\%. Notably, on the WoW dataset, the improvement approaches nearly 40\%. 

\textbf{Comparison with extractive methods.} The overall performance of extractive methods is quite satisfactory.
Fine-tuning bge-reranker with our KSE extraction yielded the best results, indicating the effectiveness of our extracted KSE. However, there still exists a discernible gap between this approach and ours, possibly highlighting the influence of the preference alignment stage and model structure.

\textbf{Comparison with abstractive methods.}
Abstractive refinement methods like Selective-Context and LongLLMLingua show a significant performance gap compared to our approach, particularly in QA tasks. This may result from their reliance on perplexity-based computations, posing a risk of losing essential entity information crucial for answering questions during refinement. In contrast, our method minimizes the risk of token-level information loss by employing sentence-level processing in data construction.

\subsection{Evaluation on Knowledge Synthesis Stage}
\begin{figure}[!ht]
    \centering
    \includegraphics[width=1.0\linewidth]{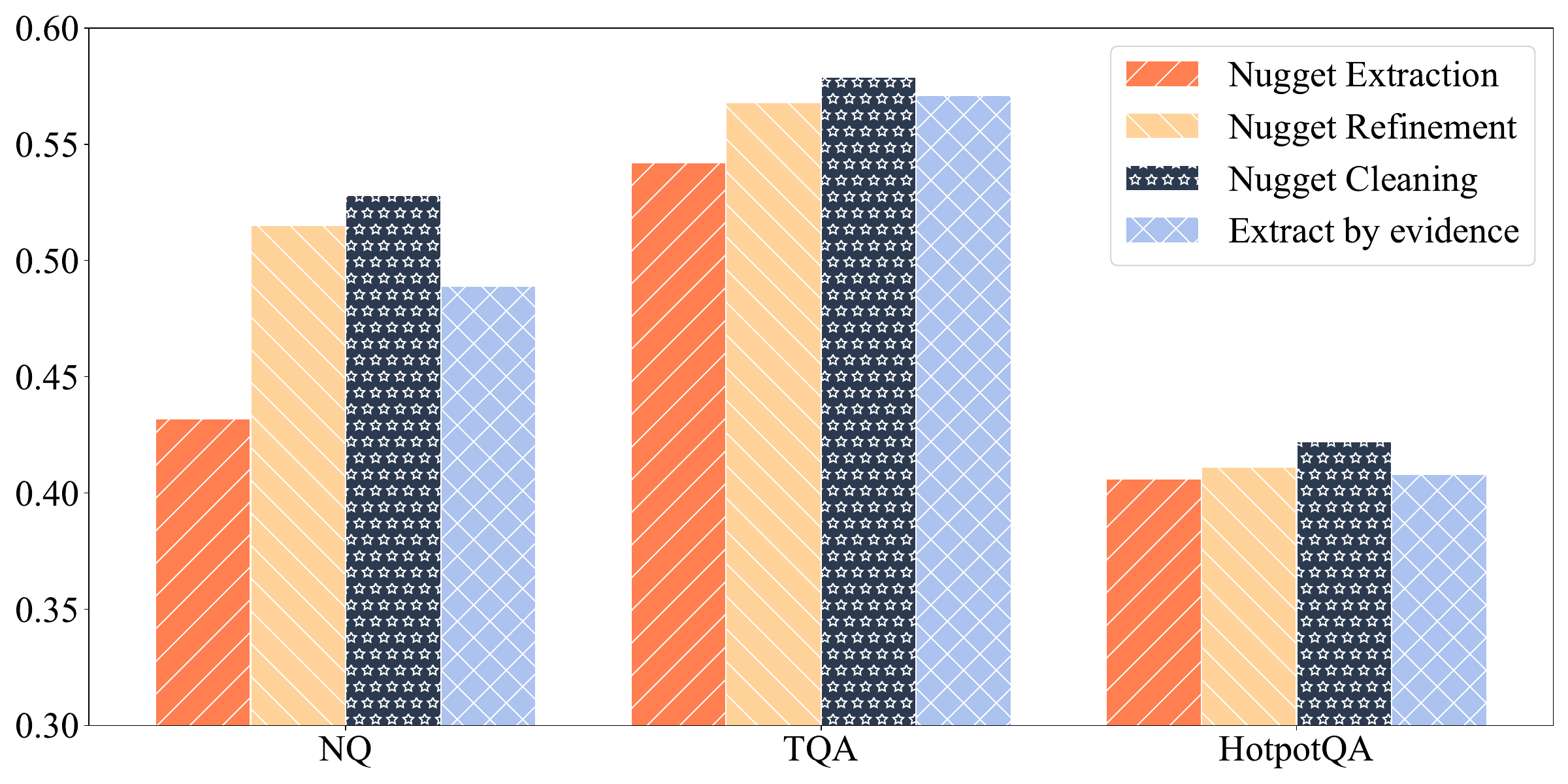}
    \caption{Performance of generator responses with different reference contents. `Nugget Extraction', `Nugget Refinement', and `Nugget Cleaning' correspond to the two intermediate products and the final output in knowledge synthesis stage, respectively.`Extract by evidence' involves extracting the top 3 sentences based on the similarity between the golden target (answer or evidence) and sentences in the retrieved documents.}
    \label{fig:semantic}
\end{figure}

To explore the necessity and effectiveness of the three steps in the knowledge synthesis stage, we use the results of each step as reference inputs for generating answers. 
For a comprehensive comparison, we incorporate results from the extraction based on the similarity between golden evidence and sentences in the retrieved documents.

As illustrated in Figure~\ref{fig:semantic}, with the further refinement of the retrieved text in the knowledge synthesis stage, the length of the input to the generator significantly decreases. However, there is a notable improvement in the quality of the LLM's responses. This observation indicates the effectiveness of our approach in reducing noise in the retrieved text, providing the generator with more easily exploitable information.
Simultaneously, it is observed that directly using golden evidence as the target for information extraction results in an inferior performance. Overall, the effectiveness is somewhat lower compared to our second step. This suggests that relying solely on the relationship between the text and the question/answer for data extraction is insufficient, and it's necessary to consider the knowledge of the model itself when constructing the data.

\subsection{Ablation Study}
\begin{table}[!t]
  \centering
  \small
    \begin{tabular}{lcccc}
          \toprule
     \multirow{2}{*}{Method}    & \multicolumn{2}{c}{NQ} & \multicolumn{2}{c}{TQA} \\
          \cmidrule(lr){2-3}\cmidrule(lr){4-5}
          & EM & \# tok & EM & \# tok \\
    \midrule
    BIDER  & 0.403 & 77    & 0.523 & 98 \\
    \quad w/o preference alig. & 0.373 & 94    & 0.518 & 85 \\
    \quad w/o knowledge syn. & 0.340  & 118   & 0.504 & 133 \\
    Original retrieval results & 0.356 & 725   & 0.480  & 787 \\
    \bottomrule
    \end{tabular}
    \caption{Ablation study on NQ and TQA.}
    \vspace{-3mm}
  \label{tab:ablation}%
\end{table}%
To assess the impact of BIDER's key components, we performed ablation experiments on NQ and TQA. Two variants were introduced for study: i) \textit{BIDER w/o preference alignment}, using models without reinforcement learning, and ii) \textit{BIDER w/o knowledge synthesis}, replacing knowledge synthesis method with a naive sentence-level retrieval method as the training target for SFT.

Table~\ref{tab:ablation} displays the results, emphasizing a decline in performance when either component is removed. This underscores the indispensability of both components.
Particularly, the impact on performance due to the absence of the knowledge synthesis method is more significant than that of the preference alignment part. This implies that the construction of the training target in the initial phase is more crucial than preference alignment. Hence, emphasizing the construction of training data in the first phase should be a priority, rather than relying solely on LLM feedback for learning.

\subsection{Impact of Preference Alignment}
\begin{table}[!t]
  \centering
  \small 
\begin{tabular}{c@{ }c@{ }c@{}c@{ }c}
\toprule
Dataset   &  Align     & \multicolumn{1}{l}{EM} & \multicolumn{1}{l}{\% Gold in Out.} & \multicolumn{1}{l}{Avg Gold Pos.} \\
\midrule
\multirow{2}[0]{*}{NQ} & \ding{53}  & 0.373 & 48.1  & 1.33 \\
      & $\checkmark$ & 0.403 & 51.1  & 1.19 \\
\midrule
\multirow{2}[0]{*}{TQA} & \ding{53} & 0.518 & 56.4  & 1.14 \\
      & $\checkmark$ & 0.523 & 61.1  & 1.10 \\
\bottomrule
\end{tabular}%
    \caption{The comparison of refiner results with and without preference alignment on NQ and TQA. `Avg Gold Pos.' represents the average position of sentences containing the golden answer (calculated only on samples that include the golden answer).}
  \label{tab:rl_comparison}
\end{table}

To further investigate the impact of preference alignment, we analyzed model output before and after this stage, specifically focusing on effective information content and its optimal sequence. We measured the proportion of golden answers in the generated results and their average position on a sentence level for models trained through supervised learning and those additionally trained with preference alignment.

Table~\ref{tab:rl_comparison} presents the results, indicating that after preference alignment training, the proportion of golden answers in the model output increased by 3\%-4\%, and their position in the output text moved closer to the beginning. This improvement suggests a dual effect: an augmentation in information content and a repositioning of crucial information towards the text's forefront.

\subsection{Impact of Retrieval Quality}
\begin{table}[!t]
  \centering
  \small
    \begin{tabular}{lcccc}
          \toprule
     \multirow{2}{*}{Method}    & \multicolumn{2}{c}{BM25} & \multicolumn{2}{c}{BM25+SimLM} \\
          \cmidrule(lr){2-3}\cmidrule(lr){4-5}
          & EM & \# tok & EM & \# tok \\
    \midrule
    Original Prompt & 0.257 & 716 & 0.356 & 725 \\
    LLM-Embedder & 0.278 & 165 & 0.357 & 161 \\
    BAET-Summarizer & 0.269 & 183 & 0.326 & 185 \\
    BIDER(ours) & 0.325 & 80 & 0.403 & 77\\
    \bottomrule
    \end{tabular}
    \caption{Experiments with different retrievers on NQ.}
    \vspace{-1mm}
  \label{tab:low_retriever}%
\end{table}%
We directly utilize top 5 retrieval documents from BM25(without reranker) to demonstrate generalizability on weaker retrievers. 
As depicted in Table~\ref{tab:low_retriever}, our approach performs well under different quality retrievers, surpassing other methods. And it can be observed that our method brings more improvements when the retriever quality is worse, indicating the effectiveness of our refinement method.

\subsection{Inference Latency}
\begin{table}[!t]
  \centering
  \small
    \begin{tabular}{lccc}
    \toprule
    Method & BIDER & Generator & Total  \\
    \midrule
    End-to-End w/o BIDER & \textbackslash{} & 1.33  & 1.33 \\
    End-to-End w/ BIDER & 0.10   & 1.08  & 1.18 \\
    \bottomrule
    \end{tabular}%
    \caption{Inference latency (seconds/query) on NQ.}
  \label{tab:latency}%
\end{table}%

Table~\ref{tab:latency} shows the inference latency of various components within the system on a V100-32G GPU. It is observed that the time required for text refinement using BIDER is notably short, facilitating effective support for applications in the RAG scenario. Additionally, as the refined input to the generator is shorter, the time taken by the generator to produce responses has also decreased. Consequently, there is a 10\% enhancement in the overall end-to-end speed.

\section{Conclusion}

We present BIDER, a method to refine retrieved documents into KSE, addressing inconsistencies between retrieved results and the knowledge needed by the generator. 
We designed a three-step process to synthesize authentic key supporting evidence to enhance the effectiveness of supervised learning, while utilizing LLM's feedback for further alignment. Through a well-structured training process, BIDER effectively provides the generator with the necessary information to answer questions based on the original retrieval text, achieving a significant improvement in answer quality while reducing input length by 80\%.

\section*{Limitations}

Our approach has some limitations. It performs less effectively in complex datasets like HotpotQA compared with NQ and TQA, suggesting that additional factors need to be considered for complex tasks.
Also, our method requires separate training for each dataset and generator, limiting its use across different tasks and generators. 
Lastly, our datasets are based solely on Wikipedia, while real-world RAG applications involve diverse sources with varied writing styles. Optimizing for this diversity may require further refinement.


\begin{thebibliography}{53}
\expandafter\ifx\csname natexlab\endcsname\relax\def\natexlab#1{#1}\fi

\bibitem[{Arefeen et~al.(2023)Arefeen, Debnath, and Chakradhar}]{arefeen2023leancontext}
Md~Adnan Arefeen, Biplob Debnath, and Srimat Chakradhar. 2023.
\newblock Leancontext: Cost-efficient domain-specific question answering using llms.
\newblock \emph{arXiv preprint arXiv:2309.00841}.

\bibitem[{Bai et~al.(2023)Bai, Lv, Zhang, Lyu, Tang, Huang, Du, Liu, Zeng, Hou, Dong, Tang, and Li}]{bai2023longbench}
Yushi Bai, Xin Lv, Jiajie Zhang, Hongchang Lyu, Jiankai Tang, Zhidian Huang, Zhengxiao Du, Xiao Liu, Aohan Zeng, Lei Hou, Yuxiao Dong, Jie Tang, and Juanzi Li. 2023.
\newblock Longbench: A bilingual, multitask benchmark for long context understanding.
\newblock \emph{arXiv preprint arXiv:2308.14508}.

\bibitem[{Bang et~al.(2023)Bang, Cahyawijaya, Lee, Dai, Su, Wilie, Lovenia, Ji, Yu, Chung, Do, Xu, and Fung}]{bang2023hallucination}
Yejin Bang, Samuel Cahyawijaya, Nayeon Lee, Wenliang Dai, Dan Su, Bryan Wilie, Holy Lovenia, Ziwei Ji, Tiezheng Yu, Willy Chung, Quyet~V. Do, Yan Xu, and Pascale Fung. 2023.
\newblock \href {http://arxiv.org/abs/2302.04023} {A multitask, multilingual, multimodal evaluation of chatgpt on reasoning, hallucination, and interactivity}.

\bibitem[{Borgeaud et~al.(2022)Borgeaud, Mensch, Hoffmann, Cai, Rutherford, Millican, Van Den~Driessche, Lespiau, Damoc, Clark et~al.}]{borgeaud2022retro}
Sebastian Borgeaud, Arthur Mensch, Jordan Hoffmann, Trevor Cai, Eliza Rutherford, Katie Millican, George~Bm Van Den~Driessche, Jean-Baptiste Lespiau, Bogdan Damoc, Aidan Clark, et~al. 2022.
\newblock Improving language models by retrieving from trillions of tokens.
\newblock In \emph{International Conference on Machine Learning}, pages 2206--2240. PMLR.

\bibitem[{Chen et~al.(2023)Chen, Pasunuru, Weston, and Celikyilmaz}]{chen2023walking}
Howard Chen, Ramakanth Pasunuru, Jason Weston, and Asli Celikyilmaz. 2023.
\newblock Walking down the memory maze: Beyond context limit through interactive reading.
\newblock \emph{arXiv preprint arXiv:2310.05029}.

\bibitem[{Dettmers et~al.(2022)Dettmers, Lewis, Belkada, and Zettlemoyer}]{NEURIPS2022_gpt3int}
Tim Dettmers, Mike Lewis, Younes Belkada, and Luke Zettlemoyer. 2022.
\newblock \href {https://proceedings.neurips.cc/paper_files/paper/2022/file/c3ba4962c05c49636d4c6206a97e9c8a-Paper-Conference.pdf} {Gpt3.int8(): 8-bit matrix multiplication for transformers at scale}.
\newblock In \emph{Advances in Neural Information Processing Systems}, volume~35, pages 30318--30332. Curran Associates, Inc.

\bibitem[{Dinan et~al.(2019)Dinan, Roller, Shuster, Fan, Auli, and Weston}]{dinan2018wizard}
Emily Dinan, Stephen Roller, Kurt Shuster, Angela Fan, Michael Auli, and Jason Weston. 2019.
\newblock \href {https://openreview.net/forum?id=r1l73iRqKm} {{W}izard of {W}ikipedia: Knowledge-powered conversational agents}.
\newblock In \emph{International Conference on Learning Representations}.

\bibitem[{Fujita et~al.(2021)Fujita, Nagarajan, Kataoka, and Ishikawa}]{PFRL}
Yasuhiro Fujita, Prabhat Nagarajan, Toshiki Kataoka, and Takahiro Ishikawa. 2021.
\newblock \href {http://jmlr.org/papers/v22/20-376.html} {Chainerrl: A deep reinforcement learning library}.
\newblock \emph{Journal of Machine Learning Research}, 22(77):1--14.

\bibitem[{Gilbert et~al.(2023)Gilbert, Sandborn, Schmidt, Spencer-Smith, and White}]{gilbert2023semantic}
Henry Gilbert, Michael Sandborn, Douglas~C Schmidt, Jesse Spencer-Smith, and Jules White. 2023.
\newblock Semantic compression with large language models.
\newblock \emph{arXiv preprint arXiv:2304.12512}.

\bibitem[{Gu et~al.(2016)Gu, Lu, Li, and Li}]{gu2016copynet}
Jiatao Gu, Zhengdong Lu, Hang Li, and Victor~OK Li. 2016.
\newblock Incorporating copying mechanism in sequence-to-sequence learning.
\newblock \emph{arXiv preprint arXiv:1603.06393}.

\bibitem[{Guu et~al.(2020)Guu, Lee, Tung, Pasupat, and Chang}]{guu2020realm}
Kelvin Guu, Kenton Lee, Zora Tung, Panupong Pasupat, and Ming-Wei Chang. 2020.
\newblock \href {https://dl.acm.org/doi/abs/10.5555/3524938.3525306} {{REALM}: Retrieval-augmented language model pre-training}.
\newblock In \emph{International Conference on Machine Learning}. JMLR.org.

\bibitem[{Izacard et~al.(2022)Izacard, Lewis, Lomeli, Hosseini, Petroni, Schick, Dwivedi-Yu, Joulin, Riedel, and Grave}]{izacard2022atlas}
Gautier Izacard, Patrick Lewis, Maria Lomeli, Lucas Hosseini, Fabio Petroni, Timo Schick, Jane Dwivedi-Yu, Armand Joulin, Sebastian Riedel, and Edouard Grave. 2022.
\newblock \href {http://arxiv.org/abs/2208.03299} {Atlas: Few-shot learning with retrieval augmented language models}.

\bibitem[{Jiang et~al.(2023{\natexlab{a}})Jiang, Wu, , Luo, Li, Lin, Yang, and Qiu}]{jiang-etal-2023-longllmlingua}
Huiqiang Jiang, Qianhui Wu, , Xufang Luo, Dongsheng Li, Chin-Yew Lin, Yuqing Yang, and Lili Qiu. 2023{\natexlab{a}}.
\newblock \href {https://arxiv.org/abs/2310.06839} {{L}ong{LLML}ingua: Accelerating and enhancing llms in long context scenarios via prompt compression}.
\newblock \emph{ArXiv preprint}, abs/2310.06839.

\bibitem[{Jiang et~al.(2023{\natexlab{b}})Jiang, Xu, Gao, Sun, Liu, Dwivedi-Yu, Yang, Callan, and Neubig}]{jiang-etal-2023-active}
Zhengbao Jiang, Frank Xu, Luyu Gao, Zhiqing Sun, Qian Liu, Jane Dwivedi-Yu, Yiming Yang, Jamie Callan, and Graham Neubig. 2023{\natexlab{b}}.
\newblock \href {https://doi.org/10.18653/v1/2023.emnlp-main.495} {Active retrieval augmented generation}.
\newblock In \emph{Proceedings of the 2023 Conference on Empirical Methods in Natural Language Processing}, pages 7969--7992, Singapore. Association for Computational Linguistics.

\bibitem[{Joshi et~al.(2017)Joshi, Choi, Weld, and Zettlemoyer}]{joshi-etal-2017-triviaqa}
Mandar Joshi, Eunsol Choi, Daniel Weld, and Luke Zettlemoyer. 2017.
\newblock \href {https://doi.org/10.18653/v1/P17-1147} {{T}rivia{QA}: A large scale distantly supervised challenge dataset for reading comprehension}.
\newblock In \emph{Proceedings of the 55th Annual Meeting of the Association for Computational Linguistics (Volume 1: Long Papers)}, pages 1601--1611, Vancouver, Canada. Association for Computational Linguistics.

\bibitem[{Karpukhin et~al.(2020)Karpukhin, Oguz, Min, Lewis, Wu, Edunov, Chen, and Yih}]{karpukhin-etal-2020-dense}
Vladimir Karpukhin, Barlas Oguz, Sewon Min, Patrick Lewis, Ledell Wu, Sergey Edunov, Danqi Chen, and Wen-tau Yih. 2020.
\newblock \href {https://doi.org/10.18653/v1/2020.emnlp-main.550} {Dense passage retrieval for open-domain question answering}.
\newblock In \emph{Proceedings of the 2020 Conference on Empirical Methods in Natural Language Processing (EMNLP)}, pages 6769--6781, Online. Association for Computational Linguistics.

\bibitem[{Kwiatkowski et~al.(2019)Kwiatkowski, Palomaki, Redfield, Collins, Parikh, Alberti, Epstein, Polosukhin, Devlin, Lee, Toutanova, Jones, Kelcey, Chang, Dai, Uszkoreit, Le, and Petrov}]{kwiatkowski-etal-2019-natural}
Tom Kwiatkowski, Jennimaria Palomaki, Olivia Redfield, Michael Collins, Ankur Parikh, Chris Alberti, Danielle Epstein, Illia Polosukhin, Jacob Devlin, Kenton Lee, Kristina Toutanova, Llion Jones, Matthew Kelcey, Ming-Wei Chang, Andrew~M. Dai, Jakob Uszkoreit, Quoc Le, and Slav Petrov. 2019.
\newblock \href {https://doi.org/10.1162/tacl_a_00276} {Natural questions: A benchmark for question answering research}.
\newblock \emph{Transactions of the Association for Computational Linguistics}, 7:452--466.

\bibitem[{Laskar et~al.(2023)Laskar, Rahman, Jahan, Hoque, and Huang}]{laskar2023cqsumdp}
Md~Tahmid~Rahman Laskar, Mizanur Rahman, Israt Jahan, Enamul Hoque, and Jimmy Huang. 2023.
\newblock Cqsumdp: A chatgpt-annotated resource for query-focused abstractive summarization based on debatepedia.
\newblock \emph{arXiv preprint arXiv:2305.06147}.

\bibitem[{Lewis et~al.(2019{\natexlab{a}})Lewis, Liu, Goyal, Ghazvininejad, Mohamed, Levy, Stoyanov, and Zettlemoyer}]{bart-cnn}
Mike Lewis, Yinhan Liu, Naman Goyal, Marjan Ghazvininejad, Abdelrahman Mohamed, Omer Levy, Veselin Stoyanov, and Luke Zettlemoyer. 2019{\natexlab{a}}.
\newblock \href {http://arxiv.org/abs/1910.13461} {{BART:} denoising sequence-to-sequence pre-training for natural language generation, translation, and comprehension}.
\newblock \emph{CoRR}, abs/1910.13461.

\bibitem[{Lewis et~al.(2019{\natexlab{b}})Lewis, Liu, Goyal, Ghazvininejad, Mohamed, Levy, Stoyanov, and Zettlemoyer}]{bart-large}
Mike Lewis, Yinhan Liu, Naman Goyal, Marjan Ghazvininejad, Abdelrahman Mohamed, Omer Levy, Veselin Stoyanov, and Luke Zettlemoyer. 2019{\natexlab{b}}.
\newblock \href {http://arxiv.org/abs/1910.13461} {{BART:} denoising sequence-to-sequence pre-training for natural language generation, translation, and comprehension}.
\newblock \emph{CoRR}, abs/1910.13461.

\bibitem[{Lewis et~al.(2020)Lewis, Perez, Piktus, Petroni, Karpukhin, Goyal, K\"{u}ttler, Lewis, Yih, Rockt\"{a}schel, Riedel, and Kiela}]{lewis2020retrieval}
Patrick Lewis, Ethan Perez, Aleksandra Piktus, Fabio Petroni, Vladimir Karpukhin, Naman Goyal, Heinrich K\"{u}ttler, Mike Lewis, Wen-tau Yih, Tim Rockt\"{a}schel, Sebastian Riedel, and Douwe Kiela. 2020.
\newblock \href {https://proceedings.neurips.cc/paper_files/paper/2020/file/6b493230205f780e1bc26945df7481e5-Paper.pdf} {{R}etrieval-{A}ugmented {G}eneration for knowledge-intensive {NLP} tasks}.
\newblock In \emph{Advances in Neural Information Processing Systems}, volume~33, pages 9459--9474.

\bibitem[{Li(2023)}]{li2023selectivecontext}
Yucheng Li. 2023.
\newblock Unlocking context constraints of llms: Enhancing context efficiency of llms with self-information-based content filtering.
\newblock \emph{arXiv preprint arXiv:2304.12102}.

\bibitem[{Lin et~al.(2021)Lin, Ma, Lin, Yang, Pradeep, and Nogueira}]{Lin_etal_SIGIR2021_Pyserini}
Jimmy Lin, Xueguang Ma, Sheng-Chieh Lin, Jheng-Hong Yang, Ronak Pradeep, and Rodrigo Nogueira. 2021.
\newblock {Pyserini}: A {Python} toolkit for reproducible information retrieval research with sparse and dense representations.
\newblock In \emph{Proceedings of the 44th Annual International ACM SIGIR Conference on Research and Development in Information Retrieval (SIGIR 2021)}, pages 2356--2362.

\bibitem[{Liu et~al.(2023)Liu, Lin, Hewitt, Paranjape, Bevilacqua, Petroni, and Liang}]{liu2023lost}
Nelson~F Liu, Kevin Lin, John Hewitt, Ashwin Paranjape, Michele Bevilacqua, Fabio Petroni, and Percy Liang. 2023.
\newblock Lost in the middle: How language models use long contexts.
\newblock \emph{arXiv preprint arXiv:2307.03172}.

\bibitem[{Liu(2019)}]{liu2019bertsum}
Yang Liu. 2019.
\newblock Fine-tune bert for extractive summarization.
\newblock \emph{arXiv preprint arXiv:1903.10318}.

\bibitem[{Loshchilov and Hutter(2019)}]{loshchilov2018adamw}
Ilya Loshchilov and Frank Hutter. 2019.
\newblock \href {https://openreview.net/forum?id=Bkg6RiCqY7} {Decoupled weight decay regularization}.
\newblock In \emph{International Conference on Learning Representations}.

\bibitem[{Mallen et~al.(2022)Mallen, Asai, Zhong, Das, Hajishirzi, and Khashabi}]{mallen2023llm_memorization}
Alex Mallen, Akari Asai, Victor Zhong, Rajarshi Das, Hannaneh Hajishirzi, and Daniel Khashabi. 2022.
\newblock When not to trust language models: Investigating effectiveness and limitations of parametric and non-parametric memories.
\newblock \emph{arXiv preprint}.

\bibitem[{OpenAI et~al.(2023)OpenAI, :, Achiam, Adler, Agarwal, Ahmad, Akkaya, Aleman, Almeida, Altenschmidt, Altman, Anadkat, Avila, Babuschkin, Balaji, Balcom, Baltescu, Bao, Bavarian, Belgum, Bello, Berdine, Bernadett-Shapiro, Berner, Bogdonoff, Boiko, Boyd, Brakman, Brockman, Brooks, Brundage, Button, Cai, Campbell, Cann, Carey, Carlson, Carmichael, Chan, Chang, Chantzis, Chen, Chen, Chen, Chen, Chen, Chess, Cho, Chu, Chung, Cummings, Currier, Dai, Decareaux, Degry, Deutsch, Deville, Dhar, Dohan, Dowling, Dunning, Ecoffet, Eleti, Eloundou, Farhi, Fedus, Felix, Fishman, Forte, Fulford, Gao, Georges, Gibson, Goel, Gogineni, Goh, Gontijo-Lopes, Gordon, Grafstein, Gray, Greene, Gross, Gu, Guo, Hallacy, Han, Harris, He, Heaton, Heidecke, Hesse, Hickey, Hickey, Hoeschele, Houghton, Hsu, Hu, Hu, Huizinga, Jain, Jain, Jang, Jiang, Jiang, Jin, Jin, Jomoto, Jonn, Jun, Kaftan, Łukasz Kaiser, Kamali, Kanitscheider, Keskar, Khan, Kilpatrick, Kim, Kim, Kim, Kirchner, Kiros, Knight, Kokotajlo, Łukasz Kondraciuk,
  Kondrich, Konstantinidis, Kosic, Krueger, Kuo, Lampe, Lan, Lee, Leike, Leung, Levy, Li, Lim, Lin, Lin, Litwin, Lopez, Lowe, Lue, Makanju, Malfacini, Manning, Markov, Markovski, Martin, Mayer, Mayne, McGrew, McKinney, McLeavey, McMillan, McNeil, Medina, Mehta, Menick, Metz, Mishchenko, Mishkin, Monaco, Morikawa, Mossing, Mu, Murati, Murk, Mély, Nair, Nakano, Nayak, Neelakantan, Ngo, Noh, Ouyang, O'Keefe, Pachocki, Paino, Palermo, Pantuliano, Parascandolo, Parish, Parparita, Passos, Pavlov, Peng, Perelman, de~Avila Belbute~Peres, Petrov, de~Oliveira~Pinto, Michael, Pokorny, Pokrass, Pong, Powell, Power, Power, Proehl, Puri, Radford, Rae, Ramesh, Raymond, Real, Rimbach, Ross, Rotsted, Roussez, Ryder, Saltarelli, Sanders, Santurkar, Sastry, Schmidt, Schnurr, Schulman, Selsam, Sheppard, Sherbakov, Shieh, Shoker, Shyam, Sidor, Sigler, Simens, Sitkin, Slama, Sohl, Sokolowsky, Song, Staudacher, Such, Summers, Sutskever, Tang, Tezak, Thompson, Tillet, Tootoonchian, Tseng, Tuggle, Turley, Tworek, Uribe, Vallone,
  Vijayvergiya, Voss, Wainwright, Wang, Wang, Wang, Ward, Wei, Weinmann, Welihinda, Welinder, Weng, Weng, Wiethoff, Willner, Winter, Wolrich, Wong, Workman, Wu, Wu, Wu, Xiao, Xu, Yoo, Yu, Yuan, Zaremba, Zellers, Zhang, Zhang, Zhao, Zheng, Zhuang, Zhuk, and Zoph}]{achiam2023gpt}
OpenAI, :, Josh Achiam, Steven Adler, Sandhini Agarwal, Lama Ahmad, Ilge Akkaya, Florencia~Leoni Aleman, Diogo Almeida, Janko Altenschmidt, Sam Altman, Shyamal Anadkat, Red Avila, Igor Babuschkin, Suchir Balaji, Valerie Balcom, Paul Baltescu, Haiming Bao, Mo~Bavarian, Jeff Belgum, Irwan Bello, Jake Berdine, Gabriel Bernadett-Shapiro, Christopher Berner, Lenny Bogdonoff, Oleg Boiko, Madelaine Boyd, Anna-Luisa Brakman, Greg Brockman, Tim Brooks, Miles Brundage, Kevin Button, Trevor Cai, Rosie Campbell, Andrew Cann, Brittany Carey, Chelsea Carlson, Rory Carmichael, Brooke Chan, Che Chang, Fotis Chantzis, Derek Chen, Sully Chen, Ruby Chen, Jason Chen, Mark Chen, Ben Chess, Chester Cho, Casey Chu, Hyung~Won Chung, Dave Cummings, Jeremiah Currier, Yunxing Dai, Cory Decareaux, Thomas Degry, Noah Deutsch, Damien Deville, Arka Dhar, David Dohan, Steve Dowling, Sheila Dunning, Adrien Ecoffet, Atty Eleti, Tyna Eloundou, David Farhi, Liam Fedus, Niko Felix, Simón~Posada Fishman, Juston Forte, Isabella Fulford, Leo Gao,
  Elie Georges, Christian Gibson, Vik Goel, Tarun Gogineni, Gabriel Goh, Rapha Gontijo-Lopes, Jonathan Gordon, Morgan Grafstein, Scott Gray, Ryan Greene, Joshua Gross, Shixiang~Shane Gu, Yufei Guo, Chris Hallacy, Jesse Han, Jeff Harris, Yuchen He, Mike Heaton, Johannes Heidecke, Chris Hesse, Alan Hickey, Wade Hickey, Peter Hoeschele, Brandon Houghton, Kenny Hsu, Shengli Hu, Xin Hu, Joost Huizinga, Shantanu Jain, Shawn Jain, Joanne Jang, Angela Jiang, Roger Jiang, Haozhun Jin, Denny Jin, Shino Jomoto, Billie Jonn, Heewoo Jun, Tomer Kaftan, Łukasz Kaiser, Ali Kamali, Ingmar Kanitscheider, Nitish~Shirish Keskar, Tabarak Khan, Logan Kilpatrick, Jong~Wook Kim, Christina Kim, Yongjik Kim, Hendrik Kirchner, Jamie Kiros, Matt Knight, Daniel Kokotajlo, Łukasz Kondraciuk, Andrew Kondrich, Aris Konstantinidis, Kyle Kosic, Gretchen Krueger, Vishal Kuo, Michael Lampe, Ikai Lan, Teddy Lee, Jan Leike, Jade Leung, Daniel Levy, Chak~Ming Li, Rachel Lim, Molly Lin, Stephanie Lin, Mateusz Litwin, Theresa Lopez, Ryan Lowe,
  Patricia Lue, Anna Makanju, Kim Malfacini, Sam Manning, Todor Markov, Yaniv Markovski, Bianca Martin, Katie Mayer, Andrew Mayne, Bob McGrew, Scott~Mayer McKinney, Christine McLeavey, Paul McMillan, Jake McNeil, David Medina, Aalok Mehta, Jacob Menick, Luke Metz, Andrey Mishchenko, Pamela Mishkin, Vinnie Monaco, Evan Morikawa, Daniel Mossing, Tong Mu, Mira Murati, Oleg Murk, David Mély, Ashvin Nair, Reiichiro Nakano, Rajeev Nayak, Arvind Neelakantan, Richard Ngo, Hyeonwoo Noh, Long Ouyang, Cullen O'Keefe, Jakub Pachocki, Alex Paino, Joe Palermo, Ashley Pantuliano, Giambattista Parascandolo, Joel Parish, Emy Parparita, Alex Passos, Mikhail Pavlov, Andrew Peng, Adam Perelman, Filipe de~Avila Belbute~Peres, Michael Petrov, Henrique~Ponde de~Oliveira~Pinto, Michael, Pokorny, Michelle Pokrass, Vitchyr Pong, Tolly Powell, Alethea Power, Boris Power, Elizabeth Proehl, Raul Puri, Alec Radford, Jack Rae, Aditya Ramesh, Cameron Raymond, Francis Real, Kendra Rimbach, Carl Ross, Bob Rotsted, Henri Roussez, Nick Ryder,
  Mario Saltarelli, Ted Sanders, Shibani Santurkar, Girish Sastry, Heather Schmidt, David Schnurr, John Schulman, Daniel Selsam, Kyla Sheppard, Toki Sherbakov, Jessica Shieh, Sarah Shoker, Pranav Shyam, Szymon Sidor, Eric Sigler, Maddie Simens, Jordan Sitkin, Katarina Slama, Ian Sohl, Benjamin Sokolowsky, Yang Song, Natalie Staudacher, Felipe~Petroski Such, Natalie Summers, Ilya Sutskever, Jie Tang, Nikolas Tezak, Madeleine Thompson, Phil Tillet, Amin Tootoonchian, Elizabeth Tseng, Preston Tuggle, Nick Turley, Jerry Tworek, Juan Felipe~Cerón Uribe, Andrea Vallone, Arun Vijayvergiya, Chelsea Voss, Carroll Wainwright, Justin~Jay Wang, Alvin Wang, Ben Wang, Jonathan Ward, Jason Wei, CJ~Weinmann, Akila Welihinda, Peter Welinder, Jiayi Weng, Lilian Weng, Matt Wiethoff, Dave Willner, Clemens Winter, Samuel Wolrich, Hannah Wong, Lauren Workman, Sherwin Wu, Jeff Wu, Michael Wu, Kai Xiao, Tao Xu, Sarah Yoo, Kevin Yu, Qiming Yuan, Wojciech Zaremba, Rowan Zellers, Chong Zhang, Marvin Zhang, Shengjia Zhao, Tianhao
  Zheng, Juntang Zhuang, William Zhuk, and Barret Zoph. 2023.
\newblock \href {http://arxiv.org/abs/2303.08774} {Gpt-4 technical report}.

\bibitem[{Ouyang et~al.(2022)Ouyang, Wu, Jiang, Almeida, Wainwright, Mishkin, Zhang, Agarwal, Slama, Ray, Schulman, Hilton, Kelton, Miller, Simens, Askell, Welinder, Christiano, Leike, and Lowe}]{NEURIPS2022_instruct}
Long Ouyang, Jeffrey Wu, Xu~Jiang, Diogo Almeida, Carroll Wainwright, Pamela Mishkin, Chong Zhang, Sandhini Agarwal, Katarina Slama, Alex Ray, John Schulman, Jacob Hilton, Fraser Kelton, Luke Miller, Maddie Simens, Amanda Askell, Peter Welinder, Paul~F Christiano, Jan Leike, and Ryan Lowe. 2022.
\newblock \href {https://proceedings.neurips.cc/paper_files/paper/2022/file/b1efde53be364a73914f58805a001731-Paper-Conference.pdf} {Training language models to follow instructions with human feedback}.
\newblock In \emph{Advances in Neural Information Processing Systems}, volume~35, pages 27730--27744. Curran Associates, Inc.

\bibitem[{Petroni et~al.(2020)Petroni, Lewis, Piktus, Rockt{\"{a}}schel, Wu, Miller, and Riedel}]{howaffects}
Fabio Petroni, Patrick S.~H. Lewis, Aleksandra Piktus, Tim Rockt{\"{a}}schel, Yuxiang Wu, Alexander~H. Miller, and Sebastian Riedel. 2020.
\newblock \href {http://arxiv.org/abs/2005.04611} {How context affects language models' factual predictions}.
\newblock \emph{CoRR}, abs/2005.04611.

\bibitem[{Petroni et~al.(2021)Petroni, Piktus, Fan, Lewis, Yazdani, De~Cao, Thorne, Jernite, Karpukhin, Maillard, Plachouras, Rockt{\"a}schel, and Riedel}]{petroni-etal-2021-kilt}
Fabio Petroni, Aleksandra Piktus, Angela Fan, Patrick Lewis, Majid Yazdani, Nicola De~Cao, James Thorne, Yacine Jernite, Vladimir Karpukhin, Jean Maillard, Vassilis Plachouras, Tim Rockt{\"a}schel, and Sebastian Riedel. 2021.
\newblock \href {https://doi.org/10.18653/v1/2021.naacl-main.200} {{KILT}: a benchmark for knowledge intensive language tasks}.
\newblock In \emph{Proceedings of the 2021 Conference of the North American Chapter of the Association for Computational Linguistics: Human Language Technologies}, pages 2523--2544, Online. Association for Computational Linguistics.

\bibitem[{Press et~al.(2023)Press, Zhang, Min, Schmidt, Smith, and Lewis}]{press-etal-2023-measuring}
Ofir Press, Muru Zhang, Sewon Min, Ludwig Schmidt, Noah Smith, and Mike Lewis. 2023.
\newblock \href {https://doi.org/10.18653/v1/2023.findings-emnlp.378} {Measuring and narrowing the compositionality gap in language models}.
\newblock In \emph{Findings of the Association for Computational Linguistics: EMNLP 2023}, pages 5687--5711, Singapore. Association for Computational Linguistics.

\bibitem[{Reimers and Gurevych(2019)}]{reimers-2019-sentence-bert}
Nils Reimers and Iryna Gurevych. 2019.
\newblock \href {https://arxiv.org/abs/1908.10084} {Sentence-bert: Sentence embeddings using siamese bert-networks}.
\newblock In \emph{Proceedings of the 2019 Conference on Empirical Methods in Natural Language Processing}. Association for Computational Linguistics.

\bibitem[{Schulman et~al.(2016)Schulman, Moritz, Levine, Jordan, and Abbeel}]{GAE}
John Schulman, Philipp Moritz, Sergey Levine, Michael~I. Jordan, and Pieter Abbeel. 2016.
\newblock \href {http://arxiv.org/abs/1506.02438} {High-dimensional continuous control using generalized advantage estimation}.
\newblock In \emph{4th International Conference on Learning Representations, {ICLR} 2016, San Juan, Puerto Rico, May 2-4, 2016, Conference Track Proceedings}.

\bibitem[{Schulman et~al.(2017)Schulman, Wolski, Dhariwal, Radford, and Klimov}]{Schulman2017PPO}
John Schulman, Filip Wolski, Prafulla Dhariwal, Alec Radford, and Oleg Klimov. 2017.
\newblock \href {https://api.semanticscholar.org/CorpusID:28695052} {Proximal policy optimization algorithms}.
\newblock \emph{ArXiv}, abs/1707.06347.

\bibitem[{See et~al.(2017)See, Liu, and Manning}]{see-etal-2017-get}
Abigail See, Peter~J. Liu, and Christopher~D. Manning. 2017.
\newblock \href {https://doi.org/10.18653/v1/P17-1099} {Get to the point: Summarization with pointer-generator networks}.
\newblock In \emph{Proceedings of the 55th Annual Meeting of the Association for Computational Linguistics (Volume 1: Long Papers)}, pages 1073--1083, Vancouver, Canada. Association for Computational Linguistics.

\bibitem[{Shi et~al.(2023{\natexlab{a}})Shi, Chen, Misra, Scales, Dohan, Chi, Schärli, and Zhou}]{shi2023irrelevant}
Freda Shi, Xinyun Chen, Kanishka Misra, Nathan Scales, David Dohan, Ed~Chi, Nathanael Schärli, and Denny Zhou. 2023{\natexlab{a}}.
\newblock \href {https://arxiv.org/pdf/2302.00093} {Large language models can be easily distracted by irrelevant context}.
\newblock \emph{arXiv preprint arXiv:2302.00093}.

\bibitem[{Shi et~al.(2023{\natexlab{b}})Shi, Min, Yasunaga, Seo, James, Lewis, Zettlemoyer, and Yih}]{shi2023replug}
Weijia Shi, Sewon Min, Michihiro Yasunaga, Minjoon Seo, Rich James, Mike Lewis, Luke Zettlemoyer, and Wen-tau Yih. 2023{\natexlab{b}}.
\newblock Replug: Retrieval-augmented black-box language models.
\newblock \emph{arXiv preprint arXiv:2301.12652}.

\bibitem[{Shinn et~al.(2023)Shinn, Cassano, Berman, Gopinath, Narasimhan, and Yao}]{shinn2023reflexion}
Noah Shinn, Federico Cassano, Edward Berman, Ashwin Gopinath, Karthik Narasimhan, and Shunyu Yao. 2023.
\newblock \href {http://arxiv.org/abs/2303.11366} {Reflexion: Language agents with verbal reinforcement learning}.

\bibitem[{Stiennon et~al.(2020)Stiennon, Ouyang, Wu, Ziegler, Lowe, Voss, Radford, Amodei, and Christiano}]{NEURIPS2020_summarizehf}
Nisan Stiennon, Long Ouyang, Jeffrey Wu, Daniel Ziegler, Ryan Lowe, Chelsea Voss, Alec Radford, Dario Amodei, and Paul~F Christiano. 2020.
\newblock \href {https://proceedings.neurips.cc/paper_files/paper/2020/file/1f89885d556929e98d3ef9b86448f951-Paper.pdf} {Learning to summarize with human feedback}.
\newblock In \emph{Advances in Neural Information Processing Systems}, volume~33, pages 3008--3021. Curran Associates, Inc.

\bibitem[{Thorne et~al.(2018)Thorne, Vlachos, Christodoulopoulos, and Mittal}]{thorne-etal-2018-fever}
James Thorne, Andreas Vlachos, Christos Christodoulopoulos, and Arpit Mittal. 2018.
\newblock \href {https://doi.org/10.18653/v1/N18-1074} {{FEVER}: a large-scale dataset for fact extraction and {VER}ification}.
\newblock In \emph{Proceedings of the 2018 Conference of the North {A}merican Chapter of the Association for Computational Linguistics: Human Language Technologies, Volume 1 (Long Papers)}, pages 809--819, New Orleans, Louisiana. Association for Computational Linguistics.

\bibitem[{Touvron et~al.(2023)Touvron, Martin, Stone, Albert, Almahairi, Babaei, Bashlykov, Batra, Bhargava, Bhosale et~al.}]{touvron2023llama}
Hugo Touvron, Louis Martin, Kevin Stone, Peter Albert, Amjad Almahairi, Yasmine Babaei, Nikolay Bashlykov, Soumya Batra, Prajjwal Bhargava, Shruti Bhosale, et~al. 2023.
\newblock \href {https://arxiv.org/pdf/2307.09288} {Llama 2: Open foundation and fine-tuned chat models}.
\newblock \emph{arXiv preprint arXiv:2307.09288}.

\bibitem[{Wang et~al.(2022)Wang, Yang, Huang, Jiao, Yang, Jiang, Majumder, and Wei}]{wang2022e5}
Liang Wang, Nan Yang, Xiaolong Huang, Binxing Jiao, Linjun Yang, Daxin Jiang, Rangan Majumder, and Furu Wei. 2022.
\newblock Text embeddings by weakly-supervised contrastive pre-training.
\newblock \emph{arXiv preprint arXiv:2212.03533}.

\bibitem[{Wang et~al.(2023{\natexlab{a}})Wang, Yang, Huang, Jiao, Yang, Jiang, Majumder, and Wei}]{wang-etal-2023-simlm}
Liang Wang, Nan Yang, Xiaolong Huang, Binxing Jiao, Linjun Yang, Daxin Jiang, Rangan Majumder, and Furu Wei. 2023{\natexlab{a}}.
\newblock \href {https://doi.org/10.18653/v1/2023.acl-long.125} {{S}im{LM}: Pre-training with representation bottleneck for dense passage retrieval}.
\newblock In \emph{Proceedings of the 61st Annual Meeting of the Association for Computational Linguistics (Volume 1: Long Papers)}, pages 2244--2258, Toronto, Canada. Association for Computational Linguistics.

\bibitem[{Wang et~al.(2023{\natexlab{b}})Wang, Li, Sun, and Liu}]{wang-etal-2023-self-knowledge}
Yile Wang, Peng Li, Maosong Sun, and Yang Liu. 2023{\natexlab{b}}.
\newblock \href {https://doi.org/10.18653/v1/2023.findings-emnlp.691} {Self-knowledge guided retrieval augmentation for large language models}.
\newblock In \emph{Findings of the Association for Computational Linguistics: EMNLP 2023}, pages 10303--10315, Singapore. Association for Computational Linguistics.

\bibitem[{Wolf et~al.(2020)Wolf, Debut, Sanh, Chaumond, Delangue, Moi, Cistac, Rault, Louf, Funtowicz, Davison, Shleifer, von Platen, Ma, Jernite, Plu, Xu, Le~Scao, Gugger, Drame, Lhoest, and Rush}]{wolf-etal-2020-transformers}
Thomas Wolf, Lysandre Debut, Victor Sanh, Julien Chaumond, Clement Delangue, Anthony Moi, Pierric Cistac, Tim Rault, Remi Louf, Morgan Funtowicz, Joe Davison, Sam Shleifer, Patrick von Platen, Clara Ma, Yacine Jernite, Julien Plu, Canwen Xu, Teven Le~Scao, Sylvain Gugger, Mariama Drame, Quentin Lhoest, and Alexander Rush. 2020.
\newblock \href {https://doi.org/10.18653/v1/2020.emnlp-demos.6} {Transformers: State-of-the-art natural language processing}.
\newblock In \emph{Proceedings of the 2020 Conference on Empirical Methods in Natural Language Processing: System Demonstrations}, pages 38--45, Online. Association for Computational Linguistics.

\bibitem[{Xiao et~al.(2023)Xiao, Liu, Zhang, and Muennighoff}]{bge_embedding}
Shitao Xiao, Zheng Liu, Peitian Zhang, and Niklas Muennighoff. 2023.
\newblock \href {http://arxiv.org/abs/2309.07597} {C-pack: Packaged resources to advance general chinese embedding}.

\bibitem[{Xu et~al.(2023)Xu, Shi, and Choi}]{xu2023recomp}
Fangyuan Xu, Weijia Shi, and Eunsol Choi. 2023.
\newblock Recomp: Improving retrieval-augmented lms with compression and selective augmentation.
\newblock \emph{arXiv preprint arXiv:2310.04408}.

\bibitem[{Yang et~al.(2023)Yang, Li, Zhang, Wang, Cheng, Li, and Xiao}]{yang-etal-2023-prca}
Haoyan Yang, Zhitao Li, Yong Zhang, Jianzong Wang, Ning Cheng, Ming Li, and Jing Xiao. 2023.
\newblock \href {https://doi.org/10.18653/v1/2023.emnlp-main.326} {{PRCA}: Fitting black-box large language models for retrieval question answering via pluggable reward-driven contextual adapter}.
\newblock In \emph{Proceedings of the 2023 Conference on Empirical Methods in Natural Language Processing}, pages 5364--5375, Singapore. Association for Computational Linguistics.

\bibitem[{Yang et~al.(2018)Yang, Qi, Zhang, Bengio, Cohen, Salakhutdinov, and Manning}]{yang-etal-2018-hotpotqa}
Zhilin Yang, Peng Qi, Saizheng Zhang, Yoshua Bengio, William Cohen, Ruslan Salakhutdinov, and Christopher~D. Manning. 2018.
\newblock \href {https://doi.org/10.18653/v1/D18-1259} {{H}otpot{QA}: A dataset for diverse, explainable multi-hop question answering}.
\newblock In \emph{Proceedings of the 2018 Conference on Empirical Methods in Natural Language Processing}, pages 2369--2380, Brussels, Belgium. Association for Computational Linguistics.

\bibitem[{Yao et~al.(2023)Yao, Zhao, Yu, Du, Shafran, Narasimhan, and Cao}]{yao2023react}
Shunyu Yao, Jeffrey Zhao, Dian Yu, Nan Du, Izhak Shafran, Karthik Narasimhan, and Yuan Cao. 2023.
\newblock {ReAct}: Synergizing reasoning and acting in language models.
\newblock In \emph{International Conference on Learning Representations}.

\bibitem[{Zhang et~al.(2023)Zhang, Xiao, Liu, Dou, and Nie}]{llm-embedder}
Peitian Zhang, Shitao Xiao, Zheng Liu, Zhicheng Dou, and Jian-Yun Nie. 2023.
\newblock \href {http://arxiv.org/abs/2310.07554} {Retrieve anything to augment large language models}.

\bibitem[{Ziegler et~al.(2019)Ziegler, Stiennon, Wu, Brown, Radford, Amodei, Christiano, and Irving}]{Ziegler2019FineTuningLM}
Daniel~M. Ziegler, Nisan Stiennon, Jeff Wu, Tom~B. Brown, Alec Radford, Dario Amodei, Paul Christiano, and Geoffrey Irving. 2019.
\newblock \href {https://api.semanticscholar.org/CorpusID:202660943} {Fine-tuning language models from human preferences}.
\newblock \emph{ArXiv}, abs/1909.08593.

\end{thebibliography}

\bibliographystyle{acl_natbib}



\end{document}